\documentclass{article}
\usepackage[accepted]{icml2020}
\usepackage{times}
\usepackage{microtype}
\usepackage{graphicx}
\usepackage{subfigure}
\usepackage{multirow}
\usepackage{booktabs} 
\usepackage{color, colortbl}
\usepackage{xfrac}
\usepackage{amssymb}
\usepackage{amsthm}
\usepackage{mathtools, cuted}
\usepackage{algorithm,algorithmic}
\usepackage{pifont}
\newtheorem{theorem}{Theorem}
\newtheorem{lemma}{Lemma}
\newtheorem*{theorem*}{Theorem}
\newtheorem*{T1}{Theorem~\ref{thm:expressivity}}
\usepackage{hyperref}
\definecolor{mydarkblue}{rgb}{0,0.08,0.45}
\definecolor{myfavblue}{rgb}{0.1176, 0.564, 1.0}
\hypersetup{
    colorlinks=true,
    linkcolor=mydarkblue,
    citecolor=mydarkblue,
    filecolor=mydarkblue,
    urlcolor=mydarkblue}
\usepackage{cleveref}

\newcommand{\mat}[1]{\mathbf{#1}}
\newcommand{\tens}[1]{\boldsymbol{\mathcal{#1}}}
\usepackage{amsfonts}
\usepackage{amsmath}

\icmltitlerunning{Tensorized Embedding Layers}

\begin{document}
\twocolumn[
\icmltitle{Tensorized Embedding Layers}

\icmlsetsymbol{equal}{*}

\begin{icmlauthorlist}
\icmlauthor{Oleksii Hrinchuk}{equal,skol,mipt}
\icmlauthor{Valentin Khrulkov}{equal,skol}
\icmlauthor{Leyla Mirvakhabova}{equal,skol}
\icmlauthor{Elena Orlova}{skol}
\icmlauthor{Ivan Oseledets}{skol,ivm}
\end{icmlauthorlist}

\icmlaffiliation{skol}{Skolkovo Institute of Science and Technology}
\icmlaffiliation{mipt}{Moscow Institute of Physics and Technology}
\icmlaffiliation{ivm}{Institute of Numerical Mathematics, RAS, Moscow, Russia}

\icmlcorrespondingauthor{Oleksii Hrinchuk}{oleksii.hrinchuk@skoltech.ru}
\icmlcorrespondingauthor{Valentin Khrulkov}{valentin.khrulkov@skoltech.ru}

\icmlkeywords{Machine Learning, ICML}

\vskip 0.3in
]









\printAffiliationsAndNotice{\icmlEqualContribution} 

\begin{abstract}
The embedding layers transforming input words into real vectors are the key components of deep neural networks used in natural language processing. However, when the vocabulary is large, the corresponding weight matrices can be enormous, which precludes their deployment in a limited resource setting. We introduce a novel way of parametrizing embedding layers based on the Tensor Train (TT) decomposition, which allows compressing the model significantly at the cost of a negligible drop or even a slight gain in performance.  We evaluate our method on a wide range of benchmarks in natural language processing and analyze the trade-off between performance and compression ratios for a wide range of architectures, from MLPs to LSTMs and Transformers.
\end{abstract}

\section{Introduction}\label{sec:intro}
Deep neural networks (DNNs) typically used in natural language processing (NLP) employ large embeddings layers, which map the input words into continuous representations and usually have the form of lookup tables. Despite such simplicity and, arguably because of it, the resulting models are cumbersome, which may cause problems in training and deploying them in a limited resource setting. Thus, the compression of large neural networks and the development of novel lightweight architectures have become essential problems in NLP research.

One way to reduce the number of parameters in the trained model is to imply a specific structure on its weight matrices (e.g., assume that they are low-rank or can be well approximated by low-rank tensor networks). Such approaches are successful at compressing the pre-trained models, but they do not facilitate the training itself. Furthermore, they usually require an additional fine-tuning stage to recover the performance of the original model.

In this paper, we introduce a new, parameter efficient embedding layer, termed TT--embedding, which can be plugged in into any model and trained end-to-end. The benefits of our compressed TT--layer are twofold. Firstly, instead of storing huge embedding matrix, we store a sequence of much smaller 2-dimensional and 3-dimensional tensors, necessary for reconstructing the required embeddings, which allows compressing the model significantly at the cost of a negligible performance drop. Secondly, the overall number of parameters can be relatively small (and constant) during the whole training stage, which allows to use larger batches or train efficiently in a case of limited resources.

To validate the efficiency of the proposed approach, we have tested it on several popular NLP tasks. In our experiments, we have observed that the standard embeddings can be replaced by TT--embeddings with the compression ratio of $1-3$ orders without any significant drop (and sometimes even with a slight gain) of the metric of interest. Specifically, we report the following compression ratios of the embedding layers: $441$ on the IMDB sentiment classification with $1\%$ absolute increase in classification accuracy; $15$ on the WMT $2014$ En--De machine translation with $0.3$ drop in the BLEU score; $3.8$ on the WikiText-103 language modeling with $1.3$ drop in test perplexity.

Additionally, we have also evaluated our algorithm on a task of binary classification with a large number of categorical features. More concretely, we applied TT--embedding to the click through rate (CTR) prediction problem, a crucial task in the field of digital advertising. Neural networks, typically used for solving this problem, while being rather elementary, include a large number of embedding layers of significant size. As a result, a majority of model parameters that represent these layers, may occupy hundreds of gigabytes of space. We show that TT--embedding not only considerably reduces the number of parameters in such models, but also sometimes improves their accuracy.


\section{Related work}\label{sec:related}

In recent years, a large body of research was devoted to compressing and speeding up various components of neural networks used in NLP tasks. \citet{joulin2016fasttext} adapted the framework of product quantization to reduce the number of parameters in linear models used for text classification. \citet{see2016compression} proposed to compress LSTM-based neural machine translation models with pruning algorithms. \citet{lobacheva2017bayesian} showed that the recurrent models could be significantly sparsified with the help of variational dropout~\citep{kingma2015variational}. \citet{cheong2019transformers} successfully compressed the Transformer architecture with the combination of pruning and quantization.

There is a plethora of prior work on compressing the embedding layers used in NLP models. \citet{chen2018learning} proposed more compact K-way D-dimensional discrete encoding scheme to replace the ``one-hot'' encoding of categorical features, such as words in NLP taks. \citet{variani2018west} introduced WEST, a compression method based on structured sparse and structured dense decomposition of the embedding matrix. \citet{chen2018groupreduce} proposed to compress the pre-trained embedding matrix by capitalizing on the power-law distribution of words and using smaller dimensionality (lower rank) for the embeddings of less frequent words. \citet{baevski2018adaptive} used similar idea in end-to-end fashion by training such structured low-rank embeddings from scratch. However, both of these methods rely on the assumption of power-law distribution of tokens and are not efficient when dealing with other popular tokenizations, such as wordpieces~\cite{schuster2012japanese,wu2016google} or BPEs~\cite{sennrich2015neural}. The effectiveness of simple low-rank factorized embeddings has been recently re-discovered by~\citet{lan2019albert}, and we refer to this method as to important baseline. Also,~\citet{lam2018word2bits} proposed a quantization algorithm for compressing word vectors, but its benefits are orthogonal to those of low-rank matrix and tensor factorizations and they can be used together, complementing each other.

Tensor methods have also been already successfully applied to neural networks compression. \citet{novikov2015tensorizing} coined the idea of reshaping weights of fully-connected layers into high-dimensional tensors and representing them in Tensor Train (TT)~\citep{oseledets2011tensor} format. This approach was later extended to convolutional~\citep{garipov2016ultimate} and recurrent~\citep{yang2017tensor, tjandra2017compressing, yu2017long} neural networks. Furthermore,~\citet{lebedev2014speeding} showed that convolutional layers could be also compressed with canonical (CP) tensor decomposition~\citep{carroll1970analysis, harshman1970foundations}. Finally, \citet{wang2018wide} compressed both fully-connected and convolutional layers with Tensor Ring decomposition~\citep{zhao2016tensor}. Recently,~\citet{ma2019tensorized} succesfully applied Block-Term Tensor Decomposition to the compression of self-attention modules in the Transformer~\citep{vaswani2017attention} architecture. In this work, we show the benefits of applying tensor machinery to the compression of embedding layers, which are an essential component of all models used in NLP.
\section{Motivation}\label{sec:motivation}

Since most of the parameters in the NLP models occupy the embedding layers, we can greatly reduce size of the entire model by compressing these layers.  Our goal is to replace the standard embedding matrix with a more compact, yet powerful and trainable, representation which would allow us to efficiently map words into vectors.

In this section, we briefly discuss our motivation of using tensorized embedding layers instead of both standard embedding layers and their low-rank factorized counterpart.

\subsection{Compression ratio perspective}

The simplest approach to compactly represent a matrix of a large size is to use the low--rank matrix factorization, which treats matrix $\mathbf{E} \in \mathbb{R}^{I \times J}$ as a product of two matrices $\mathbf{E} = \mathbf{U}\mathbf{V}^{\top}$. Here $\mathbf{U} \in \mathbb{R}^{I \times R}$ and $\mathbf{V} \in \mathbb{R}^{J \times R}$ are much ``thinner'' matrices, and $R$ is the rank hyperparameter. Note that rather than training the model with the standard embedding layer, and then trying to compress the obtained embedding, we can initially seek the embedding matrix in the described low--rank format. Then, for evaluation and training, the individual word embedding $\mathbf{E}[i,:]$ can be computed as a product $\mathbf{U}[i,:]\mathbf{V}^{\top}$ which does not require materializing the full matrix $\mathbf{E}$. This approach reduces the number of degrees of freedom in the embedding layer from $IJ$ to $(I+J)R$.

However, typically, in the NLP tasks, the embedding dimension $J$ is much smaller than the vocabulary size $I$, and obtaining significant compression ratio using low-rank matrix factorization is problematic. In order to preserve the model performance, the rank $R$ cannot be taken very small, and the compression ratio is bounded by $\frac{IJ}{(I+J)R} \leq \frac{J}{R}$, which is close to $1$ for usually full-rank embedding matrix (see Figure 1 in \cite{chen2018learning}). To overcome this bound and achieve significant compression ratio even for matrices of disproportional dimensionalities, we reshape them into multidimensional tensors and apply the \emph{Tensor Train} decomposition, which allows for more compact representation with the number of parameters falling down to logarithmic with respect to $I$.

\subsection{Softmax bottleneck perspective}
We hypothesize that such tensorized embeddings are not only superior in terms of more efficient compression, but are more theoretically justified for the usage in NLP tasks than embedding layers based on matrix factorization. Our analysis is based on \emph{softmax bottleneck} theory~\citep{yang2017breaking} and the fact that modern NLP architectures typically use the same weights for both embedding and softmax layers~\citep{press2016using,inan2016tying}.

This theory models a natural language as a collection of pairs of a context and its conditional next token distributions: $\mathcal{L} = \lbrace (c_i, P^{*}(X|c_i) \rbrace_{i=1}^{N}$, and considers parametric language models with a Softmax function operating on a context vector $\mathbf{h}_c$ and a word embedding $\mathbf{x}_i$ to define the conditional distribution $P_{\theta}(x|c)$. Given the number of context vectors $N$, the number of tokens $M$, and dimensionality of word embeddings $d$, the following three matrices are defined: $\mathbf{H}_{\theta} \in \mathbb{R}^{N \times d}$, $\mathbf{W}_{\theta} \in \mathbb{R}^{M \times d}$, $\mathbf{A}\in \mathbb{R}^{N \times M}$. The rows of these matrices correspond to
context vectors, word embeddings, and log probabilities of the true data distribution respectively. Such language model attempts to approximate $\mathbf{A}$ (up to an addition of constant matrices corresponding to a degree of freedom in Softmax) in the form 
\begin{equation}{\label{eq:softmax}}
\mathbf{A} = \mathbf{H}_{\theta} \mathbf{W}_{\theta}^{\top}.   
\end{equation}

Note that the rank of $\mathbf{H}_{\theta} \mathbf{W}_{\theta}^{\top}$ is bounded by $d$, while the matrix $\mathbf{A}$ is presumed to be a high rank matrix \citep{yang2017tensor}, which provides an upper bound on \emph{expressivity} of such models. Now, suppose that the matrix $\mathbf{W}_{\theta}$ is additionally factorized as $\mathbf{W}_{\theta} = \mathbf{U}_{\theta} \mathbf{V}_{\theta}^{\top}$ with some rank $R$. Then the rank of right-hand side of \Cref{eq:softmax} is bounded by $R$, which further reduces expressivity of such models. Contrary to this, we show that tensorized embeddings do not reduce expressivity in the softmax bottleneck sense --- while the embedding matrix is compressed it still has \emph{full matrix rank}. We provide a rigorous statement in \Cref{sec:expressivity} and verify benefits of tensorized embeddings over low-rank factorized ones empirically in \Cref{sec:experiments}.
\section{Tensor Train embedding}\label{sec:method}

In this section, we briefly introduce the necessary notation and present the algorithm for training the TT--embedding layer. Hereinafter, by $N$-way tensor $\tens{X}$ we mean a multidimensional array:
\[
\tens{X} \in \mathbb{R}^{I_1 \times I_2 \times \dots \times I_N}.
\]
with entries $\tens{X}(i_1, \ldots, i_N),$ such that $\{0 \leq i_k < I_k\}_{k=1}^N$.


\subsection{Tensor Train decomposition}{\label{sec:tt-decomp}}

A tensor $\tens{X}$ is said to be represented in the Tensor Train (TT) format~\citep{oseledets2011tensor} if each element of $\tens{X}$ can be computed as:
\begin{multline*}
\tens{X}(i_1, i_2, \dots, i_N)  = \sum_{r_1=1}^{R_1} \sum_{r_2=1}^{R_2}\dots\sum_{r_{N-1}=1}^{R_{N-1}} \\ \tens{G}^{(1)}(i_1, r_1) \tens{G}^{(2)}(r_1, i_2, r_2)\dots\tens{G}^{(N)}(r_{N-1},i_N),
\end{multline*}

where the tensors $\tens{G}^{(k)} \in \mathbb{R}^{R_{k-1} \times I_k \times R_k}$ are the so-called \textit{TT--cores} and $R_0=R_N=1$ by definition. The minimal values of $\{R_k\}_{k=1}^{N-1}$ for which the TT--decomposition exists are called \textit{TT--ranks}. Note, that the element $\tens{X}(i_1,i_2 \dots i_N)$ is effectively the product of $2$ vectors and $N-2$ matrices:
\begin{multline*}
\tens{X}(i_1, \dots, i_N) = \underbrace{\tens{G}^{(1)}[i_1,:]}_{1 \times R_1} \underbrace{\tens{G}^{(2)}[:,i_2,:]}_{R_1 \times R_2} \dots \\ \underbrace{\tens{G}^{(N-1)}[:,i_{N-1},:]}_{R_{N-2} \times R_{N-1}} \underbrace{\tens{G}^{(N)}[:,i_N]}_{R_{N-1} \times 1},
\end{multline*}
where $\tens{G}^{(k)}[:,i_k,:]$ stands for the slice (a subset of a tensor with some indices fixed) of the corresponding TT--core $\tens{G}^{(k)}$.

The number of degrees of freedom in such a decomposition can be evaluated as
$\sum_{k=1}^N R_{k-1} I_k R_k$. Thus, in the case of small ranks, the total number of parameters required to store a tensor in TT--representation is significantly smaller than $\prod_{k=1}^{N} I_k$ parameters required to store the full tensor of the corresponding size. This observation makes the application of the TT--decomposition appealing in many problems dealing with extremely large tensors.

\subsection{TT--matrix}
\label{sec:TT--mat}

Let $\mat{X} \in \mathbb{R}^{I \times J}$ be a matrix of size $I \times J$. Given two arbitrary factorizations of its dimensions into natural numbers, $I = \prod_{k=1}^N I_k$ and $J = \prod_{k=1}^N J_k$, we can reshape\footnote{by reshape we mean a column-major \texttt{reshape} command such as \texttt{numpy.reshape} in \texttt{Python}.} and transpose this matrix into an $N$-way tensor $\tens{X} \in \mathbb{R}^{I_1 J_1 \times I_2 J_2 \times \dots \times I_N J_N}$ and then apply the TT--decomposition to it, resulting in a more compact representation.

\begin{figure*}[ht]
\centering
\includegraphics[width=\textwidth]{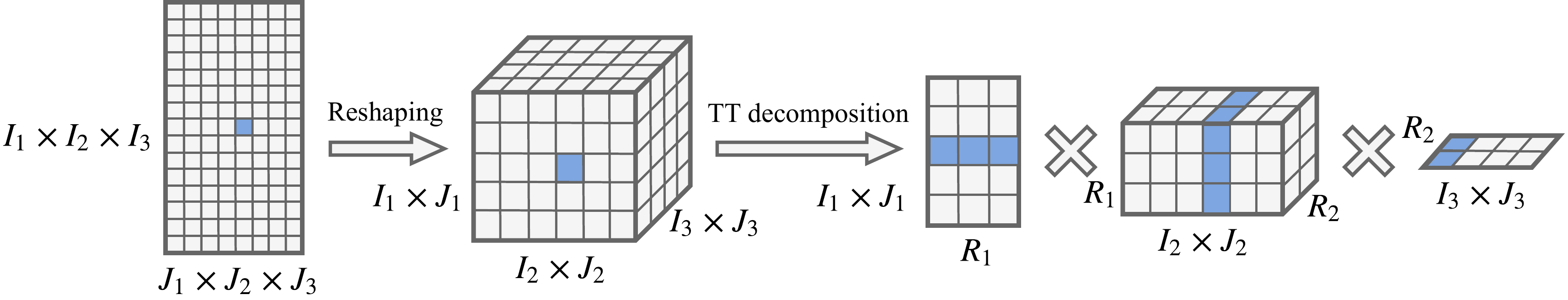}
\caption{Construction of the TT--matrix from the standard embedding matrix. Blue color depicts how the single element in the initial matrix is transformed into the product of the highlighted vectors and matrices in the TT--cores.}
\label{fig:tt_embedding}
\end{figure*}

More concretely, define the bijections $\boldsymbol{\tens{I}}(i) = (i_1,\dots,i_N)$ and $\boldsymbol{\tens{J}}(j) = (j_1,\dots,j_N)$ that map row and column indices $i$ and $j$ of the matrix $X$ to the $N$-dimensional vector-indices such that $0\leq i_k < I_k,\;0\leq j_k < J_k,\;\forall k=1,\dots,N$. From the matrix $\mat{X}$ we can form an $N$-way tensor $\tens{X}$ whose $k$-th dimension is of length $I_k J_k$ and is indexed by the tuple $(i_k,j_k)$. This tensor is then represented in the TT--format:
\begin{multline}
\label{eq:tt_matrix}
\tens{X}((i_1,j_1)\dots(i_N,j_N)) = \\ \tens{G}^{(1)}[(i_1,j_1),:]\dots\tens{G}^{(N)}[:,(i_N,j_N)].
\end{multline}
Such representation of the matrix in the TT--format is called \textit{TT--matrix}~\citep{oseledets2010approximation,novikov2015tensorizing} and is also known as Matrix Product Operator~\citep{pirvu2010matrix} in physics literature. The factorizations $(I_1, I_2, \dots I_N) \times (J_1, J_2, \dots J_N)$ will be referred to as the \emph{shape} of TT--matrix, or \textit{TT--shapes}. The process of constructing the TT--matrix from the standard matrix is visualized in~\Cref{fig:tt_embedding} for the tensor of order $3$. Note, that in this case the TT--cores are in fact $4$-th order tensors as the indices are given by tuples $(i_k,j_k)$, but all the operations defined for tensors in the TT--format are naturally extended to TT--matrices.

\subsection{TT--embedding}

By \textit{TT--embedding}, we call a layer with trainable parameters (TT--cores) represented as a TT--matrix $\tens{E}$ of the underlying tensor shape $(I_1, I_2, \dots I_N) \times (J_1, J_2, \dots J_N)$ which can be transformed into a valid embedding layer $E\in\mathbb{R}^{I \times J}$, with $I=\prod_{k=1}^{N}I_k$ and $J=\prod_{k=1}^{N}J_k$. To specify the shapes of TT--cores one has also to provide the TT--ranks, which are treated as hyperparameters of the layer and explicitly define the total compression ratio.

In order to compute the embedding for a particular word indexed $i$ in the vocabulary, we first map the row index $i$ into the $N$-dimensional vector index $(i_1,\dots,i_N)$, and then calculate components of the embedding with formula (\ref{eq:tt_matrix}). Note, that the computation of all its components is equivalent to selecting the particular slices in TT-cores (slices of shapes $J_1\times R_1$ in $\tens{G}^{(1)}$, $R_1\times J_2 \times R_2$ in $\tens{G}^{(2)}$ and so on) and performing a sequence of matrix multiplications, which is executed efficiently in modern linear algebra packages, such as BLAS. Pseudocode for the procedure of computing the mapping $i \to (i_1,\dots,i_N)$ is given in~\Cref{sec:appendix_index}.

In order to construct TT--embedding layer for a vocabulary of size $I$ and embedding dimension $J$, and to train a model with such a layer, one has to perform the following steps.
\begin{itemize}
    \item Provide factorizations of $I$ and $J$ into factors $I = I_1 \times I_2 \times \dots \times I_N$ and \mbox{$J = J_1 \times J_2 \times \dots \times J_N$}, and specify the set of TT--ranks $\lbrace R_1, R_2, \dots, R_{N-1} \rbrace$.
    
    \item Initialize the set of parameters of the embedding \mbox{$\boldsymbol{\Theta} = \lbrace \tens{G}^{(k)} \in \mathbb{R}^{R_{k-1} \times I_k \times J_k \times R_k}\rbrace_{k=1}^{N}$}. Concrete initialization scenarios are discussed further in the text.
    
    \item During training, given a batch of indices $\lbrace i_1, i_2, \dots i_b \rbrace $, compute the corresponding embeddings $\lbrace \mathbf{e}_1, \mathbf{e}_2, \dots, \mathbf{e}_b \rbrace$ using \Cref{eq:tt_matrix}.
    
    \item Computed embeddings can be followed by any standard layer such as LSTM~\citep{hochreiter1997long} or self-attention~\citep{vaswani2017attention}, and trained with backpropagation since they differentially depend on the parameters $\boldsymbol{\Theta}$.
\end{itemize}

TT--embedding implies a specific structure on the order of tokens in the vocabulary (the order of rows in the embedding matrix), and determining the optimal order is an appealing problem to solve. However, we leave this problem for future work and use the order produced by the standard tokenizer (sorted by frequency) in our current experiments.

We also experimented with a more general form of TT-decomposition, namely Tensor Ring (TR) decomposition~\citep{zhao2016tensor, wang2018wide}. This decomposition by construction has the appealing property of being circular permutation invariant (and, thus, more robust with respect to the order of the tokens), which could have potentially provided an improvement over the TT-based models with simple frequency based ordering. However, despite having stronger generalization abilities, TR might require more intricate optimization procedure (Section 2.5 in \citet{grasedyck2013literature}), and we did not observe the benefits of using TR instead of TT in our experiments (\Cref{sec:appendix_tr}).

\begin{figure}[ht!]
\centering
\includegraphics[width=0.7\linewidth]{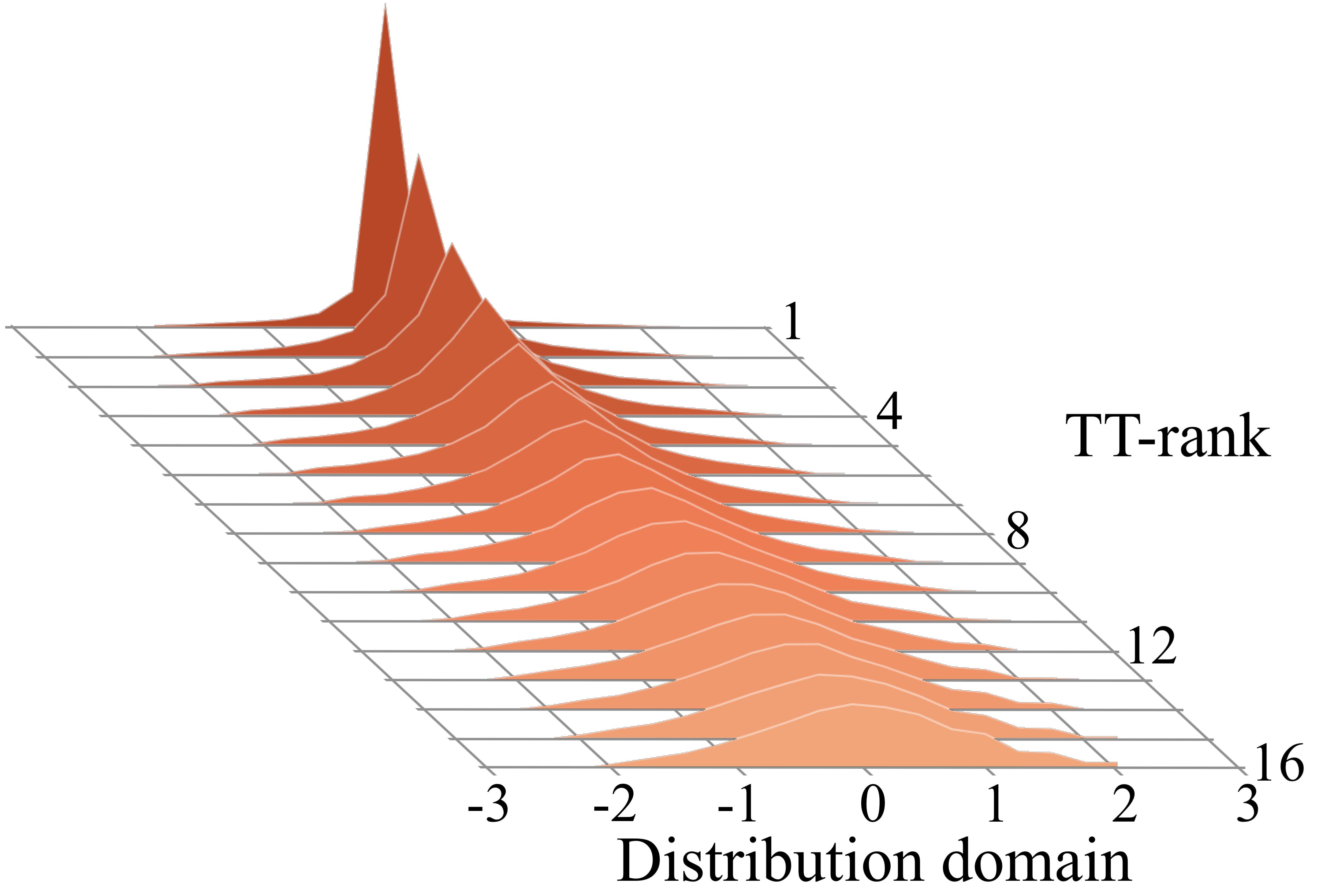}
\caption{Distribution of matrix elements of the TT--matrix of shape $(5, 5, 5, 5) \times (5, 5, 5, 5)$ initialized by formula \eqref{eq:tt_init} with $\sigma=1$. As the TT--rank increases, the resulting distribution approaches Gaussian $\mathcal{N}(0, 1)$.}
\label{fig:tt_dist}
\end{figure}

\paragraph{Initialization} The standard way to initialize an embedding matrix $\mat{E} \in \mathbb{R}^{I \times J}$ is via, e.g., Glorot initializer~\citep{glorot2010understanding}, which initializes each element as $\mat{E}(i, j) \sim \mathcal{N} \left(0, \frac{2}{I+J}\right)$. For the TT--embedding, we can only initialize the TT--cores, and the distribution of the elements of the resulting matrix $\tens{E}$ is rather non--trivial. However, it is easy to verify that if we initialize each TT--core element as $\tens{G}^{(k)}(r_{k-1},i_k, r_k) \sim \mathcal{N}(0, 1)$, the resulting distribution of the matrix elements $\tens{E}(i,j)$ has the property that $\mathbb{E}[\tens{E}(i,j)]=0$ and $\Sigma^2 \coloneqq \mathrm{Var}[\tens{E}(i,j)] =  \prod_{k=1}^{N}R_k$. Capitalizing on this observation, in order to obtain the desired variance $\mathrm{Var}[\tens{E}(i,j)]=\sigma^2$ while keeping $\mathbb{E}[\tens{E}(i,j)]=0$, we can simply initialize each TT--core as 
\begin{equation}\label{eq:tt_init}
    \tens{G}^{(k)}(r_{k-1}, i_k, r_k) \sim  \mathcal{N}\left(0, \left(\frac{\sigma}{\Sigma}\right)^{\sfrac{2}{N}}\right).
\end{equation}
The resulting distribution is not Gaussian, however, it approaches the Gaussian distribution\footnote{Asymptotic normality is a consequence of application of the Central Limit Theorem.} with the increase of the TT--rank (\Cref{fig:tt_dist}).

In our experiments, we have used the modified Glorot initializer implemented by formula \eqref{eq:tt_init}, which greatly improved performance, as opposed to initializing TT--cores simply via a standard normal distribution. It is also possible to initialize TT--embedding layer by converting the learned embedding matrix into TT--format using the TT--SVD algorithm \citep{oseledets2011tensor}, however, this approach requires the pretrained embedding matrix and does not exhibit better performance in practice~\cite{garipov2016ultimate}.

\paragraph{Hyperparameter selection} Our embedding layer introduces two additional structure-specific hyperparameters, namely \textit{TT--shapes} and \textit{TT--ranks}. 

TT--embedding does not require the vocabulary size $I$ to be represented \textit{exactly} as the product of factors $I_1, \dots, I_N$, in fact, any factorization $\prod_{k=1}^N I_k = \widetilde{I} \geq I$ will suffice. However, in order to achieve the highest possible compression ratio for a fixed value of $\widetilde{I}$, the factors $\{I_k\}_{k=1}^N$ should be as close to each other as possible~\cite{novikov2015tensorizing,yang2017tensor}. Our implementation includes a simple automated procedure for selecting a good set of values $(\{I_k\}_{k=1}^N, \{J_k\}_{k=1}^N)$ during TT--embedding initialization. The factors $J_1,\dots,J_N$ are defined by the embedding dimensionality $J$ which can be easily chosen to support good factorization, e.g., $512 = 8 \times 8 \times 8$ or $480 = 6 \times 5 \times 4 \times 4$.

The values of TT--ranks directly define the compression ratio, so choosing them to be too small or too large will result into either significant performance drop or little reduction of the number of parameters. In our experiments, we set all TT--ranks to $16$ for problems with small vocabularies and $64-192$ for problems with larger vocabularies, which resulted in a good trade-off between embedding layer compression ratio and the metric of interest.

\subsection{Expressivity of TT--embedding}{\label{sec:expressivity}}
Recall that in \Cref{sec:motivation} we argued that one advantage of TT--embeddings is the property of being full rank matrices despite providing a significant data compression. Let us now formalize this statement.

For a fixed $I=\prod_{k=1}^N I_k$, $J=\prod_{k=1}^N J_k$, and a set of ranks $\mathbf{R} = (R_1, R_2, \hdots, R_{N-1})$, we consider $\mathcal{M}_{\mathbf{R}}$, the set of all tensors represented in the TT-matrix format such that for any $\tens{X} \in \mathcal{M}_{\mathbf{R}}$ we have
\begin{equation*}
    \text{TT-rank}(\tens{X}) \leq \mathbf{R},
\end{equation*}
entry-wise. Let $\mat{X}$ denote an ordinary matrix of size $N \times M$ obtained from the TT-matrix $\tens{X}$ with the inverse of procedure decsribed in \Cref{sec:TT--mat} (application of formulas from \Cref{sec:tt-decomp}, followed by transposing and reshaping). We show that the following results holds true.
\begin{theorem}{\label{thm:expressivity}}
For all $\tens{X} \in \mathcal{M}_\mathbf{R}$ besides a set of measure zero
$$
\mathrm{rank} \ \mat{X}= \min(I, J),
$$
where the ordinary matrix rank is assumed.
\end{theorem}
See \Cref{sec:appendix_proofs} for a proof.

This theorem states that for almost all TT-embeddings (besides a negligible set), the corresponding standard embedding matrix is full-rank. Thus, using the same matrix in the softmax layer, we can achieve significant compression without hitting the softmax bottleneck, as opposed to the low-rank matrix factorization.
\section{Experiments}\label{sec:experiments}

\paragraph{Code}
We have implemented TT--embeddings described in \Cref{sec:method} in \texttt{Python} using \texttt{PyTorch} \citep{paszke2019pytorch}. 
The code is available at the anonymous repository \href{https://github.com/tt-embedding/tt-embeddings}{https://github.com/tt-embedding/tt-embeddings}.
\paragraph{Experimental setup}
We tested our approach on several popular NLP tasks:
\begin{itemize}
    \item \textbf{Sentiment analysis} --- as a starting point in our experiments, we test TT--embeddings on a rather simple task of predicting polarity of a sentence.
    \item \textbf{Neural Machine Translation (NMT)} --- to verify the applicability of TT--embeddings in more practical problems, we test it on a more challenging task of machine translation.
    \item \textbf{Language Modeling (LM)} --- then, we evaluate TT--embeddings on language modeling tasks in the case of extremely large vocabularies.
    \item \textbf{Click Through Rate (CTR) prediction} --- finally, we show that TT--embeddings can be applied for the binary classification with categorical features of significant cardinality.
\end{itemize}

\begin{table*}[htb!]
 \caption{Sentiment analysis, LSTM on IMDB and SST datasets. Embedding compression is calculated as the ratio between the number of parameters in the full embedding layer and TT--embedding layer. The LSTM parts are identical in both models, and the TT--ranks were set to $16$ in these experiments.}
\label{tab:imdb}
\begin{center}
  \begin{tabular}{ccccccc}
  \toprule
  \multirow{2}{*}{\textbf{Dataset}} & \multirow{2}{*}{\textbf{Model}} & \multirow{2}{*}{\textbf{Embedding shape}} & \multirow{2}{*}{\textbf{Test acc.}} & {\textbf{Emb}} & {\textbf{Total}} \\
   & & & & {\textbf{compr.}} & {\textbf{params}} \\
    \midrule
  \multirow{4}{*}{IMDB} & Full & $25000 \times 256$ & $0.886$ & $1$ & $7.19$M \\
  & TT1 & $(25,30,40) \times (4,8,8)$ & $0.871$ & $93$ & $0.86$M \\
   & TT2 & $(10,10,15,20) \times (4,4,4,4)$ & $0.888$ & $232$ & $0.82$M \\
   & TT3 & $(5,5,5,5,6,8) \times (2,2,2,2,4,4)$ & $\mathbf{0.897}$ & $441$ & $0.81$M \\
    \midrule
 \multirow{4}{*}{SST} &  Full  & $17200 \times 256$ & $0.374$ & $1$ & $5.19$M \\
  &  TT1  & $(24,25,30) \times (4,8,8)$ & $\mathbf{0.415}$ & $78$ & $0.85$M \\
  &  TT2  & $(10,10,12,15) \times (4,4,4,4)$ & $0.411$ & $182$ & $0.82$M \\
  &  TT3  & $(4,5,5,5,6,6) \times (2,2,2,2,4,4)$ & $0.399$ & $307$ & $0.81$M \\
    \bottomrule
  \end{tabular}
\end{center}
\end{table*}

To prove the generality and wide applicability of the proposed approach, we tested it on various architectures, such as MLPs (CTR), LSTMs (sentiment analysis), and Transformers (NMT, LM). The baselines we compare with are

\begin{enumerate}
    \item Standard embedding layer parametrized by a matrix $\mat{E}\in\mathbb{R}^{I\times J}$ with the baseline compression ratio of $1$.
    \item Low-rank factorized embedding layer parametrized by two matrices $\mat{U}\in\mathbb{R}^{I\times D}$ and $\mat{V}\in\mathbb{R}^{J\times D}$ such that the corresponding embedding matrix is $\mat{E}=\mat{U}\mat{V}^{\top}$. The compression ratio in this case is $\frac{I \times J}{(I + J) \times D} \approx \frac{J}{D}$.
\end{enumerate}

Note that Transformers in LM and NMT use the same weight matrix for their embedding and softmax layers~\citep{press2016using,inan2016tying} which already significantly reduces model size. Untying weights and tensorizing the embedding layer only will lead to the increase in the number of parameters instead of compression. In our experiments, we use two separate TT-decompositions of the same shape for embedding and softmax layers and report the compression ratios as $\frac{|V|\times d_\text{model}}{2 \times \text{TT-params}}$.

\subsection{Sentiment analysis}
\begin{table*}[htb!]
\caption{NMT, Transformer-big on WMT`14 English-to-German dataset. Both case-sensitive tokenized BLEU (higher is better) and de-tokenized SacreBLEU~\citep{post2018} on newstest2014 are reported. In case of low-rank (LR) factorization, rank is the factorization rank; in case of TT-embedding (TT), rank is the TT-rank.}
\label{tab:nmt}
\begin{center}
  \begin{tabular}{lcccccc}
    \toprule
     \multirow{2}{*}{\textbf{Model}} & \multirow{2}{*}{\textbf{Embedding shape}} & \multirow{2}{*}{\textbf{Rank}} & {\textbf{Token}} & {\textbf{Sacre}} & {\textbf{Emb}} & {\textbf{Total}} \\
     & & & {\textbf{BLEU}} & {\textbf{BLEU}} & {\textbf{compr.}} & {\textbf{params}} \\
    \midrule
    Big & $32768 \times 1024$ & --- & $\mathbf{29.58}$ & $\mathbf{28.84}$ & $1$ & $210$M \\
    \midrule
    Big+LR1 & $(32768 \times 64),\;(64 \times 1024)$ & $64$ & $28.98$ & $28.26$ & $15.5$ & $179$M \\
    Big+LR2 & $(32768 \times 32),\;(32 \times 1024)$ & $32$ & $27.79$ & $27.04$ & $31$ & $178$M \\
    Big+LR3 & $(32768 \times 16),\;(16 \times 1024)$ & $16$ & $24.80$ & $24.12$ & $62$ & $177$M \\
    \midrule
    Big+TT1 & $(32, 32, 32) \times (8, 8, 16)$ & $64$ & $29.17$ & $28.53$ & $15.3$ & $179$M \\
    Big+TT2 & $(32, 32, 32) \times (8, 8, 16)$ & $48$ & $28.53$ & $27.97$ & $26.8$ & $178$M \\
    Big+TT3 & $(32, 32, 32) \times (8, 8, 16)$ & $32$ & $28.26$ & $27.70$ & $58.5$ & $177$M \\ 
    \bottomrule
  \end{tabular}
\end{center}
\end{table*}

For this experiment, we have used the IMDB dataset~\citep{maas-EtAl:2011:ACL-HLT2011} with two categories, and the Stanford Sentiment Treebank (SST)~\citep{socher2013recursive} with five categories. We have taken the most frequent $25000$ words for the IMDB dataset and $17200$ for SST, embedded them into a $J$--dimensional space using either standard embedding or TT--embedding layer, and performed classification using a standard bidirectional two--layer LSTM with hidden size $h=128$, and dropout rate $P_\text{drop}=0.5$.

Our findings are summarized in \Cref{tab:imdb}. We observe that the models with largely compressed embedding layers can perform equally or even better than the full uncompressed models. This suggests that learning individual independent embeddings for each particular word is superfluous, as the expressive power of LSTM is sufficient to make use of these intertwined, yet more compact embeddings. Moreover, slightly better test accuracy of the compressed models in certain cases (e.g., for the SST dataset of a rather small size) insinuates that imposing specific tensorial low--rank structure on the embedding matrix can be viewed as a special form of \emph{regularization}, thus potentially improving model generalization. A detailed and comprehensive test of this hypothesis goes beyond the scope of this paper, and we leave it for future work.

\subsection{Neural Machine Translation}
For this experiment, we have trained the Transformer-big model ($d_\text{model}=1024$, $d_\text{ff}=4096$, $h=16$) from~\citet{vaswani2017attention} on WMT $2014$ English--German dataset consisting of roughly $4.5$ million sentence pairs. We evaluated on newstest2014 dataset using beam search with a beam size of $4$ and no length penalty. We did not employ checkpoint averaging and used the last checkpoint to compute the BLEU score. Sentences were tokenized with YouTokenToMe\footnote{\href{https://github.com/VKCOM/YouTokenToMe}{https://github.com/VKCOM/YouTokenToMe}} byte-pair-encodings, resulting in a joint vocabulary of $32768$ tokens. For the full list of hyperparameters, see the Appendix.

\begin{table*}[htb!]
\caption{LM, Transformer-XL~\citep{dai2018transformer} on the WikiText-103 dataset. Lower values of perplexity (PPL) are better.}
\label{tab:wikitext103}
\begin{center}
  \begin{tabular}{lcccccc}
    \toprule
     \multirow{2}{*}{\textbf{Model}} & \multirow{2}{*}{\textbf{Embedding shape}} & \multirow{2}{*}{\textbf{Rank}} & {\textbf{Valid}} & {\textbf{Test}} & {\textbf{Emb}} & {\textbf{Total}} \\
     & & & {\textbf{PPL}} & {\textbf{PPL}} & {\textbf{compr.}} & {\textbf{params}} \\
    \midrule
    TXL & $267735 \times 512$ & --- & $\mathbf{22.55}$ & $\mathbf{24.37}$ & $1$ & $192$M \\
    \midrule
    TXL+LR1 & $(267735 \times 128),\;(128 \times 512)$ & $128$ & $25.79$ & $26.92$ & $4$ & $89$M \\
    TXL+LR1 & $(267735 \times 96),\;(96 \times 512)$ & $96$ & $26.57$ & $27.75$ & $5.3$ & $81$M \\
    TXL+LR1 & $(267735 \times 48),\;(48 \times 512)$ & $64$ & $27.46$ & $28.51$ & $10.7$ & $72$M \\
    \midrule
    TXL+TT1 & $(60, 60, 75) \times (8, 8, 8)$ & $192$ & $24.38$ & $25.67$ & $3.8$ & $91$M \\
    TXL+TT2 & $(60, 60, 75) \times (8, 8, 8)$ & $128$ & $25.53$ & $26.73$ & $8.6$ & $71$M \\
    TXL+TT3 & $(60, 60, 75) \times (8, 8, 8)$ & $96$ & $26.73$ & $28.04$ & $15.1$ & $64$M \\
    \bottomrule
  \end{tabular}
\end{center}
\end{table*}

Our results are summarized in \Cref{tab:nmt}. We observe that even in this rather challenging task, both embedding and softmax layers can be compressed significantly, at the cost of a small drop in the BLEU score. However, with the increase of compression factor, the performance deteriorates rapidly. Compared to the sentiment analysis, NMT is a much more complex task which benefits more from additional capacity (in the form of more powerful RNN or more transformer blocks) rather than regularization~\citep{bahdanau2014neural,vaswani2017attention,wu2019pay}, which may explain why we did not manage to improve the model by regularizing its embedding layers with TT-embedding.

Compared to the low-rank factorization of the embedding layer, the BLEU score of the Transformer with TT-embedding is higher and degrades much slower with the decrease of TT-rank. We hypothesize that this is because of the corresponding embedding matrix being full rank and not suffering from the softmax bottleneck~\cite{yang2017breaking}.

TT-embeddings induce $8\%$ training iteration time overhead if compared to the baseline Transformer-big due to our current implementation heavy relying on slow \texttt{torch.einsum} function while standard embedding and softmax layers make use of fast and highly-optimized Tensor Cores for mixed-precision training. We expect a dedicated CUDA kernel to be much more efficient.

\subsection{Language modeling}

We took the Transformer-XL~\citep{dai2018transformer}, an open source\footnote{\href{https://github.com/kimiyoung/transformer-xl}{https://github.com/kimiyoung/transformer-xl}} state-of-the-art language modeling architecture at the time of this writing, and replaced its embedding and softmax layers with TT--factorizations. Then, we tested different model configurations on the WikiText--103~\citep{merity2016pointer} dataset and reported the results in \Cref{tab:wikitext103}. For the full list of hyperparameters, see the Appendix.

Compared to sentiment analysis and NMT, we were not able to achieve that high compression ratios for embedding and softmax layers in LM. However, in our case of extremely large vocabulary, even moderate $3.8$ times compression allowed us to save $100$M of weights at the cost of $\sim 1.5$ perplexity drop. Note that TT-embeddings also outperform low-rank factorization baseline achieving better trade-off between compression and the performance.

\begin{table*}[htb!]
\caption{CTR prediction. The hashed dataset is constructed as specified in \Cref{subsec:ctr} with hashing value $10^5$. Embedding layers with more than $2000$ unique tokens were replaced by TT--embeddings with shape factorizations consisting of $3$ or $4$ factors.}
\label{tab:ctr}
\begin{center}
  \begin{tabular}{cccccccc}
  \toprule
   \multirow{2}{*}{\textbf{Hash}} & \multirow{2}{*}{\textbf{Model}} & \multirow{2}{*}{\textbf{Factorization}} & {\textbf{TT}} & {\textbf{Hidden}} & \multicolumn{1}{c}{\textbf{Test}} & \multicolumn{1}{c}{\textbf{Emb.}} & \multicolumn{1}{c}{\textbf{Total}}\\
   & & & {\textbf{rank}} & {\textbf{size}} & \multicolumn{1}{c}{\textbf{loss}} & \multicolumn{1}{c}{\textbf{compr.}} & \multicolumn{1}{c}{\textbf{params}}\\
    \midrule
   \multirow{5}{*}{$10^5$} & Full & --- & --- & $1024$ & $0.4440$ & $1$ & $41.2$M \\
    &  TT1 & $3$ factors & $16$ & $1024$& $\mathbf{0.4433}$ & $61$ & $4.7$M \\
    & TT2 & $4$ factors & $16$ & $1024$ & $0.4440$ & $92$ & $4.5$M \\
    &  TT3 & $3$ factors & $2$ & $128$ & $0.4515$ & $2100$ & $0.53$M \\
    &  TT4 & $4$ factors & $2$ & $128$ & $0.4530$ & $4193$ & $0.53$M \\
    \midrule
    \multirow{2}{*}{---} & TT1 & $3$ factors & $16$ & $1024$ & $0.4444$ & $1004$ & $5.2$M \\
    & TT2 & $4$ factors & $16$ & $1024$ & $\mathbf{0.4438}$ & $2011$ & $4.7$M \\
    \bottomrule
  \end{tabular}
\end{center}
\end{table*}

\subsection{Click Through Rate prediction}\label{subsec:ctr}
Among other applications of the TT--embedding layer, we chose to focus on CTR prediction, a popular task in digital advertising~\citep{he2014practical}. We consider open dataset provided by Criteo for Kaggle Display Advertising Challenge~\citep{criteo} which consists of $39$ categorical features, $45.8$M samples and is binary labeled according to whether the user clicked on the given advertisement. Unique values of categorical features are bijectively mapped into integers. To reduce the memory footprint, if the size of a corresponding vocabulary is immense (e.g., a cardinality of some features in this dataset is of order $10^6$), these integers are further hashed by taking modulus with respect to some fixed number such as $10^5$. However, due to strong compression properties of TT--embeddings, this is not necessary for our approach, and we consider both full and hashed datasets in our experiments.

\paragraph{CTR with the baseline algorithm}
 The task at hand can be treated as a binary classification problem. As a baseline algorithm, we consider the neural network with the following architecture. First, each of the categorical features is passed through a separate embedding layer with embedding size $J$. After that, the embedded features are concatenated and passed through $4$ fully-connected layers of $1024$ neurons and ReLU activation functions. In all experiments, we used Adam optimizer with the learning rate equal to $0.0005$. Since many input features have a large number of unique values (e.g., $10131227$) and storing the corresponding embedding matrices would be costly, we employ the hashing procedure mentioned earlier.

\paragraph{CTR with TT--embeddings}
We substitute the embedding layers with the TT--embedding layers. Besides that, we leave the overall structure of the neural network unchanged with the same parameters as in the baseline approach. \Cref{tab:ctr} presents the experimental results on the Criteo CTR dataset. To the best of our knowledge, our loss value is very close to the state-of-the-art result~\citep{juan2016field}. These experiments indicate that the substitution of large embedding layers with TT--embeddings leads to significant compression ratios (up to $2011$ times) with a slight improvement in the test loss, and up to $4200$ with a small drop in the test loss. The total size of the compressed model does not exceed $20$ Mb, while the baseline model weighs about $160$ Mb. The obtained compression ratio suggests that the usage of TT--embedding layers may be beneficial in CTR prediction.

\section{Discussion and future work}\label{sec:discussion}

We propose a novel embedding layer, the TT--embedding, for compressing huge lookup tables used for encoding categorical features of significant cardinality, such as the index of a token in natural language processing tasks. The proposed approach, based on the TT--decomposition, experimentally proved to be effective, as it heavily decreases the number of training parameters at the cost of a small deterioration in performance. In addition, our method can be easily integrated into any deep learning framework and trained via backpropagation, while capitalizing on reduced memory requirements and increased training batch size.

Our experimental results suggest several appealing directions for future work. First of all, TT--embeddings impose a concrete tensorial low-rank structure on the embedding matrix, which was shown to improve the generalization ability of the networks acting as a regularizer. The properties and conditions of applicability of this regularizer are subject to more rigorous analysis. Secondly, unlike standard embedding, we can introduce non-linearity into TT-cores to improve their expressive power~\citep{khrulkov2019generalized}. Additionally, it is important to understand how the order of tokens in the vocabulary affects the properties of the networks with TT--embedding. We hypothesize that there exists the optimal order of tokens which better exploits the particular structure of TT--embedding and leads to a boost in performance and/or compression ratio. Finally, the idea of applying higher--order tensor decompositions to reduce the number of parameters in neural nets is complementary to more traditional methods such as pruning~\citep{han2015learning} and quantization~\citep{hubara2016quantized, xu2018deep}. Thus, it would be interesting to make a thorough comparison of all these methods and investigate whether their combination may lead to even stronger compression.

\clearpage
\bibliography{main}
\bibliographystyle{icml2020}
\clearpage
\appendix

\section{Multiindex construction}{\label{sec:appendix_index}}

\begin{algorithm}[htb!]
\caption{The algorithm implementing the bijection  $\boldsymbol{\tens{I}}(i)$ as described in \Cref{sec:TT--mat}.}
\label{alg:index-quant}
\begin{algorithmic}
\STATE {\bfseries Require}: $I$ -- vocabulary size, $\lbrace I_k \rbrace_{k=1}^{N}$ -- an arbitrary factorization of $I$,
\STATE \quad\quad\quad\quad $i$ -- index of the target word in vocabulary.
\STATE {\bfseries Returns}: $\boldsymbol{\tens{I}}(i) = (i_1,\dots,i_{N})$ -- $N$-dimensional index.
\STATE {\bfseries Initialize}: $L = \lbrace 1, I_1, I_1 I_2, \dots,I_1 I_2\dots I_{N-1}\rbrace$
\FOR{$k=N$ {\bfseries to} $1$}
\STATE $i_k \leftarrow \texttt{floor}(\sfrac{i}{ L[k]})$
\STATE $i \leftarrow i \mod L[k]$
\ENDFOR 
\end{algorithmic}
\end{algorithm}

\begin{algorithm}[htb!]
\caption{The algorithm implementing the bijection $(i_1,\dots,i_N) \rightarrow i$, inverse to $\boldsymbol{\tens{I}}(i)$.}
\label{alg:index-back}
\begin{algorithmic}
\STATE {\bfseries Require}: $I$ -- vocabulary size, $\lbrace I_k \rbrace_{k=1}^{N}$ -- an arbitrary factorization of $I$, 
\STATE \quad\quad\quad\quad $(i_1,\dots,i_{N})$ -- $N$-dimensional index.
\STATE {\bfseries Returns}: $i$ -- index of the target word in vocabulary
\STATE {\bfseries Initialize}: $L = \lbrace 1, I_1, I_1 I_2, \dots,I_1 I_2\dots I_{N-1}\rbrace$
\STATE {$i \leftarrow 0$}
\FOR{$k=1$ {\bfseries to} $N$}
\STATE $i \leftarrow i + i_k \times L[k]$
\ENDFOR 
\end{algorithmic}
\end{algorithm}
\section{Proof of \Cref{thm:expressivity}}{\label{sec:appendix_proofs}}
Recall that for fixed $I=\prod_{k=1}^N I_k$, $J=\prod_{k=1}^N J_k$, and a set of ranks $\mathbf{R} = (R_1, R_2, \hdots, R_{N-1})$ we defined $\mathcal{M}_{\mathbf{R}}$, the set of all tensors represented in the TT-matrix format such that for any $\tens{X} \in \mathcal{M}_{\mathbf{R}}$ we have
\begin{equation*}
    \text{TT-rank}(\tens{X}) \leq \mathbf{R},
\end{equation*}
entry-wise. Let $\mat{X}$ denote an ordinary matrix of size $N \times M$ obtained from the TT-matrix $\tens{X}$ with the inverse of procedure decsribed in \Cref{sec:TT--mat} (application of formulas from \Cref{sec:tt-decomp}, followed by transposing and reshaping).

Our analysis is based on the fact that $\mathcal{M}_{\mathbf{R}}$ forms an \emph{irreducible algebraic set} \citep{buczynska2015hackbusch,hartshorne2013algebraic}. Concretely, we will use the fact that for an irreducible algebraic set $\mathcal{A}$ any algebraic subset $\mathcal{B}$ either has measure zero, or coincides with $\mathcal{A}$. We start with a simple lemma.
\begin{lemma}{\label{lemma:algebraic}}
Let $$\mathcal{B} = \lbrace \tens{X} \in \mathcal{M}_{\mathbf{R}} \colon \mathrm{rank} \ \mat{X} < \min(I, J)\rbrace,$$
then $\mathcal{B}$ is an algebraic subset of $\mathcal{M}_{\mathbf{R}}$.
\end{lemma}
\begin{proof}
We need to show that $\mathcal{B}$ is cut out by polynomial equations on $\mathcal{M}_{\mathbf{R}}$. This readily follows from the facts that $\mathrm{mat}(\cdot)$ is a linear mapping, and that the upper bound on matrix rank can be specified by requiring all minors of specific size to vanish (which is a polynomial constraint).
\end{proof}
We now show that $\mathcal{B}$ is in fact a proper subset of $\mathcal{M}_{\mathbf{R}}$, i.e., $\mathcal{B} \subsetneq \mathcal{M}_{\mathbf{R}}$.

\begin{lemma}{\label{lemma:example}}
For any $\mathcal{M}_{\mathbf{R}}$ there exists $\tens{X} \in \mathcal{M}_{\mathbf{R}}$ with
$$ \mathrm{rank} \ \mat{X} = \min(I, J). $$
\end{lemma}
\begin{proof}
We provide a concrete example of such a tensor. Define the collection of TT--cores \mbox{$\lbrace \tens{G}^{(k)} \in \mathbb{R}^{1 \times I_k \times J_k \times 1} \rbrace_{k=1}^{N}$} using the equations
\begin{equation}
\tens{G}^{(k)}[:, (i, j), :] = \delta_{ij},
\end{equation}
with $\delta_{ij}$ denoting the Kronecker delta symbol. It easy to verify that $\mat{X}$ of a tensor $\tens{X}$ specified by this collection of cores takes a very simple form:
$\mat{X}[i, j] = \delta_{ij}$, which clearly is of maximal rank.
\end{proof}
Using \Cref{lemma:algebraic,lemma:example} and based on previous discussion on properties of algebraic sets we conclude that the following theorem holds.
\begin{T1}
For all $\tens{X} \in \mathcal{M}_\mathbf{R}$ besides a set of measure zero
$$
\mathrm{rank} \ \mat{X} = \min(I, J),
$$
where the ordinary matrix rank is assumed.
\end{T1}

\section{Tensor Ring Embedding}{\label{sec:appendix_tr}}

\begin{figure*}[ht]
\centering
\includegraphics[width=0.95\textwidth]{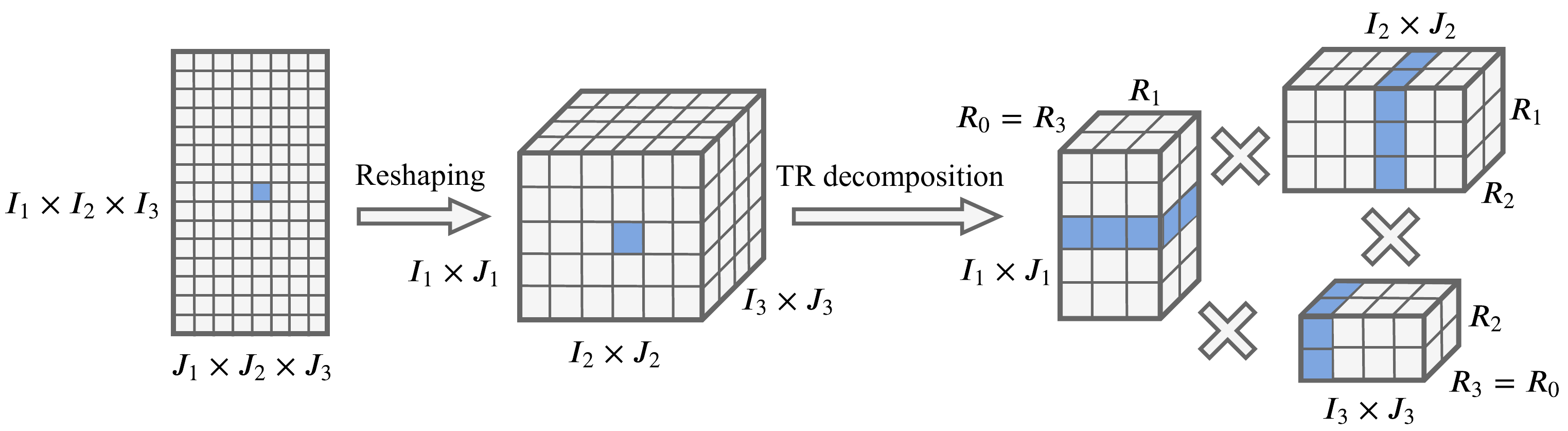}
\caption{Construction of the TR--matrix from the standard embedding matrix. Blue color depicts how the single element in the initial matrix is transformed into the product of the highlighted matrices. In contrast to TT-embedding, \textit{matrix trace} operator is applied to the final matrix, resulting in a scalar (highlighted element).}
\label{fig:tr_embedding}
\end{figure*}

Tensor Ring (TR) decomposition is a generalization to TT-decomposition where the first and the last cores are $3$-dimensional tensors which corresponds to $R_0=R_N>1$. Formally, a tensor $\tens{X}$ is said to be represented in the TR format~\citep{zhao2016tensor} if each element of $\tens{X}$ can be computed as:
\begin{multline*}
\tens{X}(i_1, i_2, \dots, i_d)  = \sum_{r_0=1}^{R_0}\sum_{r_1=1}^{R_1} \dots\sum_{r_{N-1}=1}^{R_{N-1}} \tens{G}^{(1)}(r_0,i_1, r_1) \\ \tens{G}^{(2)}(r_1, i_2, r_2)\dots\tens{G}^{(N)}(r_{N-1},i_N,r_0).
\end{multline*}
Similar to TT, we can define TR-matrix (see \Cref{fig:tr_embedding}) and corresponding TR-embedding layer.


While our results (\Cref{tab:imdb_appendix} and \Cref{tab:nmt_appendix}) suggest that TT-embedding shows better compression-performance trade-off than its TR counterpart, much more experimentation is needed to properly compare these two approaches (for example, we see that TR is a promising direction for future work as it outperforms TT on SST-2 benchmark). However, such analysis is computationally heavy and goes beyond the scope of this paper.


\begin{table*}[htb!]
 \caption{Sentiment analysis, LSTM with either TT-embedding or TR-embedding on IMDB and SST datasets.}
\label{tab:imdb_appendix}
\begin{center}
  \begin{tabular}{cccccccc}
  \toprule
  \multirow{2}{*}{\textbf{Dataset}} & \multirow{2}{*}{\textbf{Model}} & \multirow{2}{*}{\textbf{Embedding shape}} & \multirow{2}{*}{\textbf{Rank}} & \multirow{2}{*}{\textbf{Test acc.}} & {\textbf{Emb}} & {\textbf{Total}} \\
   & & & & & {\textbf{compr.}} & {\textbf{params}} \\
    \midrule
  \multirow{10}{*}{IMDB} & Full & $25000 \times 256$ & --- & $0.886$ & $1$ & $7.19$M \\
  \cmidrule{2-7}
  & TT1 & $(25,30,40) \times (4,8,8)$ & 16 & $0.871$ & $93$ & $0.86$M \\
   & TT2 & $(10,10,15,20) \times (4,4,4,4)$ & 16 & $0.888$ & $232$ & $0.82$M \\
   & TT3 & $(5,5,5,5,6,8) \times (2,2,2,2,4,4)$ & 16 & $\mathbf{0.897}$ & $441$ & $0.81$M \\
   \cmidrule{2-7}
  & TR1 & $(25,30,40) \times (4,8,8)$ & 16 & $0.869$ & $45$ & $0.87$M \\
   & TR2 & $(10,10,15,20) \times (4,4,4,4)$ & 16 & $0.872$ & $109$ & $0.86$M \\
   & TR3 & $(5,5,5,5,6,8) \times (2,2,2,2,4,4)$ & 16 & $0.884$ & $215$ & $0.82$M \\
  & TR4 & $(25,30,40) \times (4,8,8)$ & 8 & $0.854$ & $152$ & $0.83$M \\
   & TR5 & $(10,10,15,20) \times (4,4,4,4)$ & 8 & $0.882$ & $455$ & $0.80$M \\
   & TR6 & $(5,5,5,5,6,8) \times (2,2,2,2,4,4)$ & 8 & $0.890$ & $1042$ & $0.80$M \\
    \midrule[0.8pt]
 \multirow{7}{*}{SST} &  Full  & $17200 \times 256$ & --- & $0.374$ & $1$ & $5.19$M \\
 \cmidrule{2-7}
  &  TT1  & $(24,25,30) \times (4,8,8)$ & 16 & $0.415$ & $78$ & $0.85$M \\
  &  TT2  & $(10,10,12,15) \times (4,4,4,4)$ & 16 & $0.411$ & $182$ & $0.82$M \\
  &  TT3  & $(4,5,5,5,6,6) \times (2,2,2,2,4,4)$ & 16 & $0.399$ & $307$ & $0.81$M \\
 \cmidrule{2-7}
  &  TR1  & $(24,25,30) \times (4,8,8)$ & 8 & $\mathbf{0.427}$ & $128$ & $0.83$M \\
  &  TR2  & $(10,10,12,15) \times (4,4,4,4)$ & 8 & $0.411$ & $366$ & $0.80$M \\
  &  TR3  & $(4,5,5,5,6,6) \times (2,2,2,2,4,4)$ & 8 & $0.394$ & $800$ & $0.78$M \\
    \bottomrule
  \end{tabular}
\end{center}
\end{table*}

\begin{table*}[htb!]
\caption{NMT, Transformer-big with either TT-embedding or TR-embedding on WMT`14 English-to-German dataset. Both case-sensitive tokenized BLEU and de-tokenized SacreBLEU~\citep{post2018} on newstest2014 are reported.}
\label{tab:nmt_appendix}
\begin{center}
  \begin{tabular}{lccccccc}
    \toprule
     \multirow{2}{*}{\textbf{Model}} & \multirow{2}{*}{\textbf{Embedding shape}} & \multirow{2}{*}{\textbf{Rank}} & {\textbf{Token}} & {\textbf{Sacre}} & {\textbf{Emb}} & {\textbf{Total}} \\
     & & & {\textbf{BLEU}} & {\textbf{BLEU}} & {\textbf{compr.}} & {\textbf{params}} \\
    \midrule
    Big & $32768 \times 1024$ & --- & $\mathbf{29.58}$ & $\mathbf{28.84}$ & $1$ & $210$M \\
    \midrule
    Big+TT1 & $(32, 32, 32) \times (8, 8, 16)$ & $64$ & $29.17$ & $28.53$ & $15.3$ & $179$M \\
    Big+TT2 & $(32, 32, 32) \times (8, 8, 16)$ & $48$ & $28.53$ & $27.97$ & $26.8$ & $178$M \\
    Big+TT3 & $(32, 32, 32) \times (8, 8, 16)$ & $32$ & $28.26$ & $27.70$ & $58.5$ & $177$M \\
    \midrule
    Big+TR1 & $(32, 32, 32) \times (8, 8, 16)$ & $32$ & $28.64$ & $28.07$ & $16$ & $179$M \\
    Big+TR2 & $(32, 32, 32) \times (8, 8, 16)$ & $16$ & $28.10$ & $27.50$ & $64$ & $177$M \\
    \bottomrule
  \end{tabular}
\end{center}
\end{table*}

\begin{table*}[htb!]

\parbox{.45\linewidth}{
\label{tab:nmt_params}
\caption{Hyperparameters of Transformer-big used for neural machine translation on WMT`14.}
\vspace{0.3em}
  \begin{tabular}{l|l}
    \toprule
    {\textbf{Parameter}} & {\textbf{Value}} \\
    \midrule
    \textit{Data cleaning} & \\
    \quad max training sequence length in tokens & $128$ \\
    \quad max source / target ratio & $2.5$ \\
    \midrule
    \textit{Model} & \\
    \quad vocabulary size, $|V|$ & $32768$ \\
    \quad hidden size, $d_\text{model}$ & $1024$ \\
    \quad intermediate FF layer size, $d_\text{ff}$ & $4096$ \\
    \quad number of attention heads, $h$ & $16$ \\
    \quad number of layers in encoder / decoder & $6$ \\
    \midrule
    \textit{Optimization} & \\
    \quad optimizer & NovoGrad \\
    \quad learning rate & $0.04$ \\
    \quad betas, $(\beta_1,\beta_2)$ & $(0.95,0.25)$ \\
    \quad learning rate decay policy & cosine \\
    \quad weight decay & $0.0001$ \\
    \quad batch size in tokens & $393216$ \\
    \quad number of training steps & $80000$ \\
    \quad number of warmup steps & $4000$ \\
    \midrule
    \textit{Regularization} & \\
    \quad global dropout, $P_\text{drop}$ & $0.2$ \\
    \quad label smoothing & $0.1$ \\
    \midrule
    \textit{Inference} & \\
    \quad beam search beam size & $4$ \\
    \quad length penalty & $0$ \\
    \bottomrule
  \end{tabular}
}
\hfill
\parbox{.45\linewidth}{
\label{tab:lm_params}
\caption{Hyperparameters of Transformer-XL used for language modeling on WikiText-103.}
\vspace{0.3em}
  \begin{tabular}{l|l}
    \toprule
    {\textbf{Parameter}} & {\textbf{Value}} \\
    \midrule
    \textit{Model} & \\
    \quad vocabulary size, $|V|$ & $267735$ \\
    \quad hidden size, $d_\text{model}$ & $512$ \\
    \quad intermediate FF layer size, $d_\text{ff}$ & $2048$ \\
    \quad number of attention heads, $h$ & $8$ \\
    \quad number of layers & $16$ \\
    \midrule
    \textit{Optimization} & \\
    \quad optimizer & NovoGrad \\
    \quad learning rate & $0.025$ \\
    \quad betas, $(\beta_1,\beta_2)$ & $(0.95,0.25)$ \\
    \quad learning rate decay policy & cosine \\
    \quad weight decay & $0.0001$ \\
    \quad batch size in sequences & $1024$ \\
    \quad target sequence length & $128$ \\
    \quad memory sequence length & $128$ \\
    \quad number of training steps & $300000$ \\
    \quad number of warmup steps & $3000$ \\
    \midrule
    \textit{Regularization} & \\
    \quad global dropout, $P_\text{drop}$ & $0.15$ \\
    \midrule
    \textit{Inference} & \\
    \quad batch size & $4$ \\
    \quad target sequence length & $128$ \\
    \quad memory sequence length & $640$ \\
    \quad max positional encodings length & $400$ \\
    \bottomrule
  \end{tabular}
}
\end{table*}

\end{document}